\documentclass[letterpaper,times]{sae}

\usepackage[utf8]{inputenc}
\usepackage[svgnames, table, xcdraw]{xcolor}
\usepackage[justification=justified, singlelinecheck=true]{caption}
\usepackage{amsmath}
\usepackage{amssymb}
\usepackage{array}
\usepackage{booktabs}
\usepackage{calc}
\usepackage{comment}
\usepackage{float}
\usepackage{graphicx}
\usepackage{hyperref}
\usepackage{lineno}
\usepackage{multirow}
\usepackage{multimedia}
\usepackage{nameref}
\usepackage{pgfplots}
\usepackage{pgfplotstable}
\usepackage{tabularx}
\usepackage{tikz}
\usepackage{todonotes}
\usepackage{url}
\usepackage{dblfloatfix}
\usepackage{multirow}
\usepackage{booktabs}
\usepackage{array}
\usepackage{graphicx}
\usepackage[normalem]{ulem}
\usepackage{geometry}  
\usepackage{todonotes} 
\usepackage{authblk}
\hypersetup{hidelinks}
\usepackage{listings}
\usepackage{caption}

\usepackage{tikz}
\usepackage{xcolor}
\usepackage{pifont}      
\usepackage{caption}     

\definecolor{success}{RGB}{176,175,80}
\definecolor{failure}{RGB}{244,67,54}

\definecolor{SAEblue}{RGB}{1,160,233}
\definecolor{success}{RGB}{46,125,50}    
\definecolor{failure}{RGB}{198,40,40}     

\definecolor{promptLightGray}{HTML}{F5F5F5}
\definecolor{promptBorderGray}{HTML}{B0B0B0}

\lstdefinestyle{promptstyle}{
  basicstyle=\ttfamily\scriptsize,      
  breaklines=true,                 
  breakindent=8pt,                 
  breakatwhitespace=false,
  backgroundcolor=\color{promptLightGray},
  frame=single,                    
  framerule=0.8pt,                 
  rulecolor=\color{promptBorderGray},    
  showspaces=false,
  showstringspaces=false,
  showtabs=false,
  xleftmargin=0.8pt,               
  xrightmargin=0.8pt,              
  resetmargins=true,               
  breakautoindent=false,           
  columns=flexible,                
  keepspaces=true,                 
}

\DeclareCaptionFont{eightpt}{\fontsize{8pt}{11pt}\selectfont #1}
\usepgfplotslibrary{polar}
\usetikzlibrary{positioning}
\usetikzlibrary{shapes.geometric, arrows.meta, calc}

\pgfplotsset{compat=1.18}
\usetikzlibrary{patterns}

\newcolumntype{L}[1]{>{\raggedright\let\newline\\\arraybackslash\hspace{0pt}}p{#1}}
\newcolumntype{C}[1]{>{\centering\let\newline\\\arraybackslash\hspace{0pt}}p{#1}}
\newcolumntype{R}[1]{>{\raggedleft\let\newline\\\arraybackslash\hspace{0pt}}p{#1}}

\makeatletter
\def\@seccntformat#1{%
  \expandafter\csname c@#1\endcsname\c@section
  }
\makeatother

\makeatletter 
\renewcommand\@biblabel[1]{#1. } 
\makeatother

\SAECopyright{2025}

\PaperTitle{Understanding Adversarial Transferability in Vision-Language Models for Autonomous Driving: A Cross-Architecture Analysis}

\AddAuthor{David Fernandez, Pedram MohajerAnsari, Amir Salarpour and Mert D. Pesé}{School of Computing, Clemson University, USA\\ {\{dferna3, pmohaje, asalarp, mpese\}}@clemson.edu}

\begin{document}
\maketitle

\section{Abstract}
Vision-language models (VLMs) are increasingly used in autonomous driving because they combine visual perception with language-based reasoning, supporting more interpretable decision-making, yet their robustness to physical adversarial attacks, especially whether such attacks transfer across different VLM architectures, is not well understood and poses a practical risk when attackers do not know which model a vehicle uses. We address this gap with a systematic cross-architecture study of adversarial transferability in VLM-based driving, evaluating three representative architectures (Dolphins, OmniDrive, and LeapVAD) using physically realizable patches placed on roadside infrastructure in both crosswalk and highway scenarios. Our transfer-matrix evaluation shows high cross-architecture effectiveness, with transfer rates of 73--91\% (mean TR = 0.815 for crosswalk and 0.833 for highway) and sustained frame-level manipulation over 64.7--79.4\% of the critical decision window even when patches are not optimized for the target model. We further find asymmetric architecture-level risk, with Dolphins most vulnerable to incoming transfer attacks (VS = 0.82) and LeapVAD producing the most transferable patches (TO = 0.882), while models sharing CLIP-based vision encoders exhibit stronger bidirectional transfer. Overall, these results indicate that current VLM-based autonomous driving systems share systematic cross-architecture weaknesses that architectural diversity alone does not resolve, underscoring the need for defenses and design principles that explicitly account for transferability in safety-critical deployment.

\section{1. Introduction}
\label{sec:1-introduction}

Vision-language models (VLMs) are rapidly changing the field of autonomous driving (AD). By combining visual perception with language-based reasoning, these models create decision-making systems that are easier for humans to interpret~\cite{VLMsurvey}. Unlike traditional modular methods that separate perception, prediction, and planning, VLMs use large language models (LLMs) trained on massive amounts of data. This allows them to understand complex driving situations through natural language~\cite{hu2023planning, Hagedorn2023}. Recent VLMs like Dolphins~\cite{dolphins}, OmniDrive~\cite{wang2025omnidrive}, and LeapVAD~\cite{ma2025leapvad} show that VLMs can make human-like driving decisions and explain the reasoning behind their actions. These systems also show promise in handling unexpected driving scenarios, which makes them a strong alternative to older AD designs.


\begin{figure}[!t]
    \centering
    \includegraphics[width=\columnwidth]{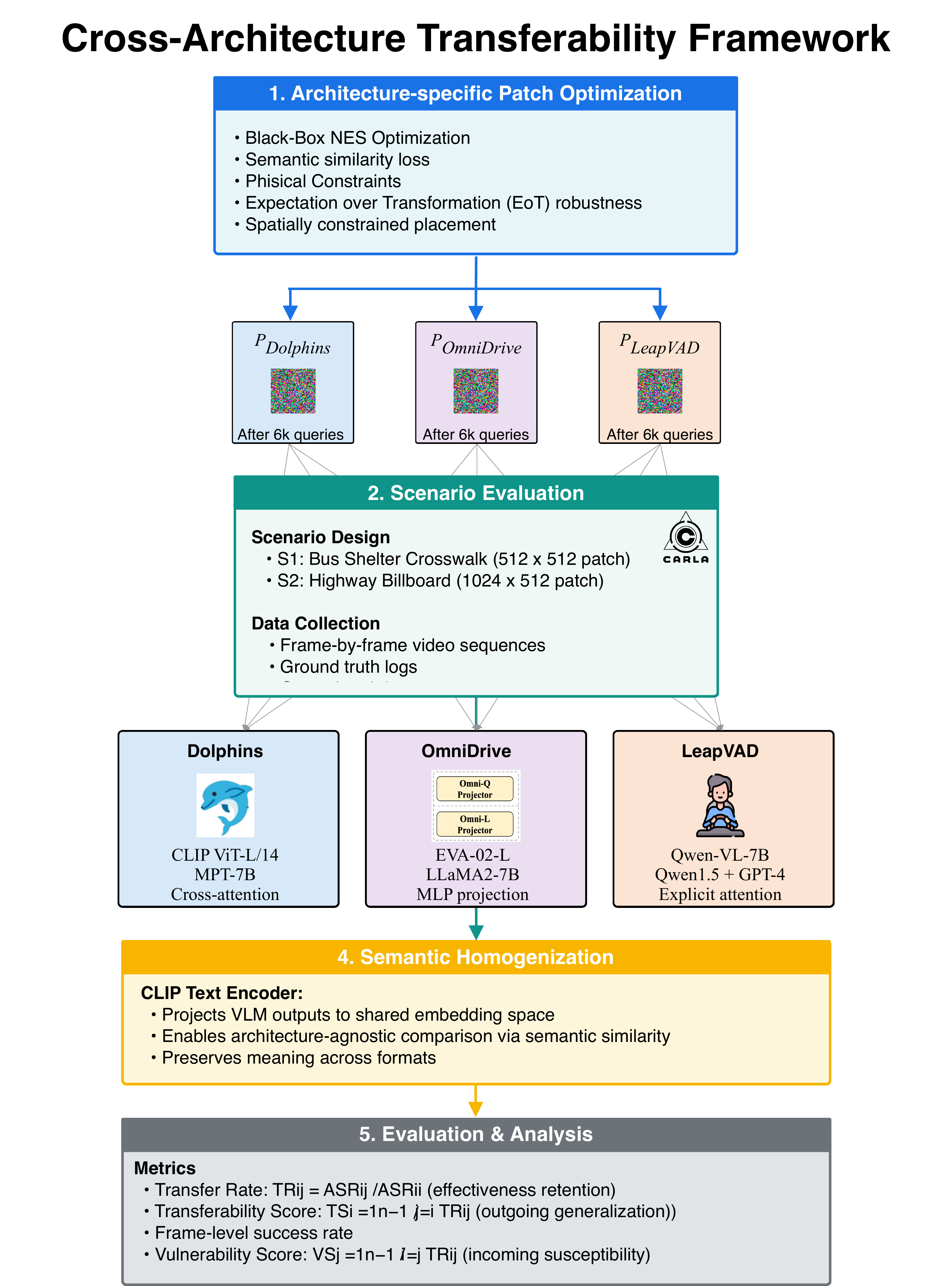}
    \caption{\textbf{Cross-Architecture Adversarial Transferability Framework.} Five-stage pipeline to evaluate how adversarial patches transfer across VLM architectures: (1) generate patches (three architecture-specific and one universal), (2) record scenarios in CARLA, (3) test all 9 patch–VLM pairings and build a transfer matrix, (4) normalize model outputs using a CLIP text encoder for consistent scoring, and (5) analyze transferability patterns.}
    \label{fig:framework}
\end{figure}

Despite their potential, the robustness of VLM-based driving systems against physical adversarial attacks, such as adversarial patches, remains a major concern~\cite{eykholt2018robust}. Physical adversarial patches are specially designed visual patterns that can be printed and placed in the driving environment to deceive AI systems~\cite{wei2024physical}. While researchers have studied these attacks on traditional computer vision models~\cite{adversarialPatchBrown}, the unique architecture of VLMs, which tightly couples visual encoders with LLMs through cross-modal fusion mechanisms, creates a fundamentally different and more complex attack surface. A major unanswered question is whether these attacks can transfer from one VLM architecture to another. This is a vital issue since there are now over 20 different VLM designs~\cite{vlmList} using various vision and language components. If a single patch placed on a roadside sign can trick vehicles from many different manufacturers, the risk is much higher. 


This paper presents a systematic analysis of adversarial attack transferability across VLM-based AD architectures. We establish a comparative framework to address three core research questions regarding the extent of attack transferability (RQ1), the robustness of specific architectural designs (RQ2), and if the attacks maintain frame-level success with patches that target other architectures(RQ3). To answer these questions, we develop a transfer matrix methodology to evaluate success rates across 12 model combinations using three VLM architectures: Dolphins, which uses cross-attention fusion; OmniDrive, which employs MLP projection; and LeapVAD, which utilizes a dual-process system with explicit attention. These models were selected to represent diverse vision-language architectures while maintaining comparability through shared camera-only sensing and open-source availability. Our method generates four distinct adversarial patches: three tailored specifically to each architecture and one universal patch optimized across all models using multi-objective black-box optimization. Finally, we conduct closed-loop testing in the CARLA~\cite{CARLA} simulation environment, focusing on scenarios that target both pedestrian detection and vehicle steering logic.


Our evaluation reveals that adversarial patches transfer well across architectures, with transfer rates of 0.73--0.91, meaning patches optimized for one model remain effective on others. Transfer is slightly higher on highways (mean TR = 0.833) than at crosswalks (mean TR = 0.815), consistent with the idea that simpler visual settings can make transfer easier. We observe meaningful differences across architectures. Dolphins is most exposed to incoming transfer attacks (VS = 0.82), while OmniDrive shows partial isolation due to its EVA-CLIP encoder (VS = 0.810), although it remains substantially vulnerable. LeapVAD produces the most transferable patches (TO = 0.882), suggesting its attacks exploit shared weaknesses in the visual feature space. Consistent with this, the CLIP-based models (Dolphins and LeapVAD) cluster in transfer behavior, indicating that shared vision encoders are a key driver of cross-architecture vulnerability. Frame-level results show these attacks are also persistent: self-attacks manipulate 78.5--89.1\% of frames, and transferred attacks still succeed on 64.7--79.4\% of frames, indicating sustained influence during most of the critical decision window.

These findings highlight a serious security gap in current VLM-based autonomous driving systems. High cross-architecture transferability (73--91\%) and sustained persistence over time (64--79\% frame success in transfer settings) show that an attacker can launch effective attacks without knowing the vehicle’s specific VLM implementation.

This paper makes the following contributions:
\begin{itemize}
    \item A reproducible transfer matrix approach to evaluate adversarial robustness across multiple architectures. By measuring success rates across 12 distinct model-patch combinations, we demonstrate that transferability is a vital metric for VLM security
    \item A black-box optimization method using Natural Evolution Strategies, to create universal patches capable of deceiving various VLM architectures simultaneously.
    \item Our evaluation shows that adversarial transferability in VLM-based driving is a systematic issue, with transfer rates ranging from 73\% to 91\%.
\end{itemize}

\section{2. Related Work}
\label{sec:2-related_work}

\subsection{2.1. End-to-End Vision-Language Models for Autonomous Driving}

Recent advances in VLMs have led to end-to-end driving systems that combine perception and decision-making into a single architecture. Models like Dolphins~\cite{lingo}, LINGO-1~\cite{lingo}, and DriveGPT4~\cite{drivegpt} use LLMs with visual encoders to process driving scenes~\cite{fernandezWIP, fernandez2026forensic}, generating both natural language reasoning~\cite{fernandez2025avoiding} and control actions. These systems typically employ pre-trained vision encoders (like CLIP ViT~\cite{li2022blip}) linked to LLMs through cross-modal fusion mechanisms such as cross-attention~\cite{dolphins}, MLP projection~\cite{wang2025omnidrive}, or explicit attention~\cite{ma2025leapvad}. This integration allows the model to connect visual observations to its reasoning about traffic rules and safety. Unlike traditional systems that use separate modules for perception, prediction, and planning~\cite{hu2023planning}, VLM-based systems promise to better handle corner cases by leveraging the broad world knowledge learned during pre-training on internet-scale datasets. While individual VLM architectures have demonstrated promising performance in normal driving scenarios, their robustness to adversarial attacks and the extent to which vulnerabilities transfer across different architectural designs remain unexplored. 


\subsection{2.2. Adversarial Attacks on Computer Vision}
Physical adversarial patches have proven to be a significant vulnerability for computer vision systems. Brown et al.~\cite{adversarialPatchBrown} showed that printed patterns could reliably fool image classifiers in the real world. Later research extended these attacks to object detectors~\cite{eykholt2018robust, mohajeransari2025attention} and segmentation models~\cite{segmentation}, revealing that safety-critical perception components can be fooled by these carefully designed visual patterns. Eykholt et al.~\cite{eykholt2018robust} demonstrated attacks on traffic sign recognition using simple stickers, highlighting a direct threat to autonomous vehicle perception. A comprehensive survey by Wei et al.~\cite{wei2024physical} documents a decade of progress in physical adversarial attacks, establishing that real-world deployment of deep learning systems faces persistent security challenges from adversarial manipulation.
A key challenge for these physical attacks is making them robust to real-world viewing changes. Expectation over Transformation (EoT)~\cite{ExpectationOverTransformation} is a method designed to address this challenge by optimizing attacks to remain effective across a range of viewing angles, distances, and environmental conditions rather than a single perfect setup. Our work adapts these established techniques to the more complex domain of VLM-based autonomous driving. In our case, the attacks must work over time (across video frames), must influence both what the VLM perceives and how it reasons, and must transfer across architectures with heterogeneous vision-language integration mechanisms.

%

\subsection{2.3. Adversarial Robustness of Multimodal Models}
Recent work has started to examine the adversarial vulnerabilities of multimodal VLMs. Studies on models like CLIP~\cite{CLIP}, BLIP~\cite{li2022blip}, and LLaVA~\cite{liu2023llava} show that these combined architectures have unique weaknesses that vision-only systems do not exhibit. Attacks can exploit the vision-language connection to cause specific erroneous outputs~\cite{VLATTACK}, manipulating both visual encoders and cross-modal attention mechanisms. However, most of this research focuses on general-purpose VLMs using single images for tasks like image captioning or visual Q\&A, rather than safety-critical sequential decision-making.

Our work extends this research to the safety-critical domain of AD with several key differences. First, we target video-based, sequential decision-making in closed-loop driving scenarios, not single-image classification tasks. Second, we impose a temporal consistency requirement, since driving systems must be reliable frame after frame to ensure safety. Third, we focus on physically realizable patches placed in realistic infrastructure locations (bus shelters, billboards), rather than unrestricted digital perturbations. Most importantly, while prior work examines vulnerabilities of individual VLM architectures in isolation, we investigate whether adversarial attacks transfer across architectures, a critical question for deployment where attackers may not know which specific VLM a target vehicle employs. These specific constraints make our threat model more realistic and challenging, as patches must work under the changing viewing conditions of a moving vehicle while remaining effective across heterogeneous VLM designs.


\subsection{2.4 Universal Adversarial Perturbations}

Universal adversarial perturbations represent a particularly severe threat to deployed machine learning systems, as they demonstrate that single adversarial patterns can fool models across different inputs without requiring input-specific optimization. Moosavi-Dezfooli et al.~\cite{MoosaviAdversarial} introduced the concept of universal adversarial perturbations for image classifiers, showing that a single perturbation applied to any input image could cause misclassification with high probability. This work revealed that decision boundaries of deep networks share common geometric properties that can be exploited universally. Subsequent research extended universal perturbations to object detection~\cite{LiUniversalAdvPerturb}, semantic segmentation~\cite{metzen2017universalAdvSemantic}, and other computer vision tasks, demonstrating that universality is a fundamental property of adversarial examples rather than an artifact of specific architectures.

Despite the established threat of universal perturbations in computer vision, no prior work has evaluated universal adversarial attacks against VLM-based AD systems. This gap is critical because real-world adversaries placing physical patches on infrastructure may not know which specific VLM architecture a target vehicle employs.
\section{3. Background}
\label{sec:3-background}

\subsection{3.1. VLMs Architecture}
To enable cross-architecture transferability analysis, we select three representative VLM architectures that cover different ways of integrating vision and language for AD, while remaining comparable through shared camera-only sensing and open-source availability. Each architecture fuses visual perception with language-based reasoning in a distinct way, which can lead to different vulnerability patterns. ~\autoref{tab:vlm_architecture_comparison} summarizes the key features of the three systems and highlights differences in vision encoders, language models, fusion mechanisms, and output formats that motivate our transferability study.

\begin{table*}[t]
\centering
\caption{VLM architecture comparison showing diversity in vision encoders, fusion mechanisms, and language models across the three evaluated systems.}
\label{tab:vlm_architecture_comparison}
{\small
\begin{tabular}{@{}lccccc@{}}
\toprule
\textbf{Model} & \textbf{Vision Encoder} & \textbf{Language Model} & \textbf{Fusion Mechanism} & \textbf{Parameters} & \textbf{Output Format} \\
\midrule
Dolphins & CLIP ViT-L/14 & MPT-7B & Cross-attention & 7.5B & Free-form NL \\
OmniDrive (Omni-L) & EVA-02-L & LLaMA2-7B & MLP projection & 7.6B & Structured NL + 3D \\
LeapVAD & Qwen-VL-7B & Qwen1.5-1.8B + GPT-4 & Explicit attention & 8.8B + API & Dual-process NL \\
\bottomrule
\end{tabular}
}
\end{table*}



\subsection{3.2. Black-Box Adversarial Optimization}
We used NES (Natural Evolutionary Strategies)~\cite{ilyas2018black}, a gradient-free optimization algorithm suited for black-box attacks where model gradients are unavailable. NES works by sampling the objective function (attack success) at multiple points in the parameter space to estimate a gradient, rather than using backpropagation. This approach is suitable for attacking VLMs for two reasons. First, production systems are typically black-box, this means an attacker cannot access the model’s internal architecture or weights. Interaction is limited to sending inputs (queries) and receiving outputs. Second, NES can optimize our objective even with non-differentiable steps like parsing text outputs.

\subsection{3.3. Physical Adversarial Patches}
Physical adversarial patches have unique constraints not found in digital attacks. They must be printable, which limits their color range, and must be deployed in the real world, which restricts their size and location. Furthermore, these patches must be robust to environmental changes like varying illumination, different viewing angles, and potential occlusions. To ensure this robustness, we used EoT, which optimizes the patch to remain effective across a range of these real-world variations. Our threat model also enhances realism by constraining patches to existing advertising infrastructure, such as bus shelter panels and billboards. This reflects a plausible scenario where an attacker compromises a legitimate display. This spatial constraint makes the attack more challenging, as the patch cannot be placed in an arbitrary "best" location but must succeed from a fixed position.
\section{4. Threat Model}
Our threat model assumes an adversary can interact with VLM-based AD systems in a black-box setting by submitting images and observing the outputs (such as scene descriptions and action recommendations). However, the adversary cannot determine which VLM architecture a target vehicle uses and has no access to internal model parameters, gradients, training data, or other implementation details. This reflects practical deployment conditions where attackers may gain limited insight through indirect interaction, such as querying publicly available models, observing vehicle behavior, or using restricted API endpoints during reconnaissance, but cannot identify whether a particular vehicle relies on a specific model, since manufacturers typically do not disclose these details and different fleets often use different systems.

The attacker’s goal is to trigger unsafe driving actions, such as accelerating toward pedestrians or steeritng toward barriers, by placing physically realistic adversarial patches on roadside advertising infrastructure like bus shelter panels and highway billboards. This setup matches plausible real-world attack routes, including compromising digital displays through social engineering or insider access, or printing durable, weather-resistant patches for physical installation. The attacker may scout high-traffic locations, design patches offline using accessible open-source VLMs as stand-ins, and then deploy them in the environment. However, the attacker cannot alter vehicle sensors, onboard software, or other components, and is limited to visual manipulation of the scene.



\section{5. Research Questions}
\label{sec:4-problem}

Our study of adversarial transferability in VLM-based autonomous driving focuses on three core research questions that understand how attacks carry over across different model architectures and what this means for real-world deployment choices.
\subsubsection{RQ1: Cross-Architecture Transferability}
\textbf{How well do adversarial patches optimized for one VLM architecture transfer to other architectures, and what drives these transfer rates?}

This research question examines whether adversarial weaknesses in VLM-based autonomous driving are mostly tied to specific architectures or reflect broader issues shared across different model designs. We study how patches optimized for Dolphins, OmniDrive, and LeapVAD perform when evaluated on the other architectures, and we characterize the resulting transfer patterns. We also analyze factors that may influence transfer, including the vision encoder, the fusion mechanism, and correlations in learned visual features.

\subsubsection{RQ2: Architecture-Specific Vulnerabilities}
\textbf{Which VLM architectures are more robust or more vulnerable to transfer attacks, and which design choices explain these differences?}

This question focuses on identifying which VLM designs are most susceptible to cross-architecture attacks. We measure how often each architecture is successfully attacked by patches optimized for other models and compare their overall vulnerability levels. We then relate these outcomes to key design choices, including the vision encoder, fusion mechanism, and language model size. We also examine whether high vulnerability under direct (single-model) attacks predicts high vulnerability to transfer attacks, or whether these are largely independent. The results provide practical guidance for selecting VLM architectures in security-critical autonomous driving settings.

\subsubsection{RQ3: Temporal Consistency}
\textbf{Do patches optimized for one architecture maintain sustained effectiveness when transferred to different VLMs?}

We evaluate whether adversarial patches cause brief, isolated errors or sustain influence across an entire driving sequence. Many AD safety designs assume perception mistakes are short-lived and can be handled through temporal filtering, redundancy checks, or fallback behaviors. We test whether that assumption holds by measuring frame-level success rates, defined as the fraction of frames in which the patch steers the VLM output toward the intended adversarial outcome. We compare self-attacks (patches evaluated on their source architecture) with cross-architecture transfers to see whether transferred patches remain consistently effective or become intermittent. We also examine how frame-level performance varies across source–target pairs to assess how predictable temporal persistence is in transfer settings. Together, these results indicate whether architectural diversity and temporal filtering can meaningfully reduce risk, or whether sustained manipulation can overwhelm safety mechanisms during critical moments such as crosswalk interactions.

\section{6. Methodology}
\label{sec:6-Methodology}

In this section, we detail the methodology used to analyze the transferability of adversarial attacks across VLM-based AD architectures. 

\subsection{6.1. Experimental Setup}
All experiments are conducted in CARLA 0.9.14~\cite{CARLA}. We use Town04, which offers diverse road geometries, including urban intersections and highway segments, to evaluate different attack vectors. Our experimental setup uses a single forward-facing RGB camera mounted on the ego vehicle's dashboard at 1920×1080 resolution, matching typical autonomous vehicle sensor configurations and ensuring that all VLMs process identical visual inputs for fair comparison.We evaluate adversarial transferability using two complementary scenarios: 

\textbf{Crosswalk Scenario}. The ego vehicle approaches an urban intersection at 30 km/h as a pedestrian crosses in a marked crosswalk. A 512$\times$512 adversarial patch is placed on a bus shelter advertisement on the left roadside, becoming visible at about 30 m and covering roughly 5--7\% of the image width at the key decision point (10 m from the crosswalk). In this scenario, the patch tries to reduce attention to the pedestrian, leading to unsafe recommendations such as “accelerate” or “maintain speed” instead of “brake.”

\textbf{Highway Scenario}. The ego vehicle travels at 85 km/h in the right lane of a three-lane highway with a concrete barrier along the right edge. A 1024$\times$512 adversarial patch is placed on a roadside billboard and becomes visible from about 80 m. In this scenario, the patch can trigger unsafe “turn right” recommendations toward the barrier, even when the barrier remains visible. Safe actions are “maintain speed” or “accelerate,” making any rightward turn unsafe.

\subsection{6.2. Adversarial Transferability Analysis}
Adversarial transferability describes how often attacks designed for one model still succeed against other models. This is important because an attacker placing physical patches in the environment is unlikely to know which VLM architecture a vehicle uses. High transferability suggests these weaknesses are broad and not tied to a single design, while low transferability implies that using diverse architectures may offer some natural protection.
We formalize transferability through the \textit{transfer rate} metric, which normalizes cross-model attack success relative to same-model performance:





\subsection{6.3 Adversarial Patch Generation}
We generate architecture-specific adversarial patches using Natural Evolution Strategies (NES)~\cite{naturalEvolutionStrategies}, a gradient-free black-box method that does not require access to model internals. Each patch is initialized as random noise within the RGB range $[0,255]$. At iteration $t$, NES samples $N=20$ random search directions ${\epsilon_i}_{i=1}^{N} \sim \mathcal{N}(0,\sigma^2 I)$ with $\sigma=0.1$ and evaluates two candidate patches, $\theta_t+\sigma\epsilon_i$ and $\theta_t-\sigma\epsilon_i$. We then update the patch using the resulting estimate with learning rate $\alpha=0.02$. We run 150 iterations, which corresponds to roughly 6,000 model queries per patch.

\subsubsection{6.3.1. Loss Function} 
Because the three VLMs produce different output formats, we use a unified semantic loss based on CLIP text embeddings:
\begin{equation}
L_{\text{semantic}} = 1 - \frac{e_{\text{generated}} \cdot e_{\text{target}}}{|e_{\text{generated}}| |e_{\text{target}}|}
\end{equation}
where $e_{\text{generated}}$ is the embedding of the model output and $e_{\text{target}}$ is the embedding of a target adversarial response. For the crosswalk scenario, the target is \textit{"The driver should accelerate and continue forward because the road is clear."} For the highway scenario, the target is \textit{"The driver should turn right to exit the highway."} This encourages both incorrect actions and misleading justifications.

\textit{Physical Realizability.} To keep patches deployable, we hard-clip pixel values to $[0,255]$ and add weak total-variation regularization ($\lambda_{\text{TV}}=0.001$) to promote smooth, printable patterns.

\subsubsection{6.3.2. Expectation over Transformation (EoT)} 
To improve robustness under realistic viewing changes, we optimize under random perturbations: spatial jitter ($\pm 5$ pixels), brightness scaling ($[0.9,1.1]$), and contrast shifts ($[-0.05,0.05]$). We approximate the expected loss using $K=5$ transformed samples per candidate patch:
\begin{equation}
\mathbb{E}{T \sim \mathcal{T}}[L(\theta, T)] \approx \frac{1}{K} \sum{k=1}^{K} L(\theta, T_k)
\end{equation}

\subsubsection{6.3.3. Physical Realizability Constraints} 
To ensure that optimized patches could be printed and deployed in real-world scenarios, we enforced several constraints during optimization. All RGB pixel values were hard-clipped to range $[0, 255]$ at each iteration, guaranteeing compatibility with standard color printers and displays. Finally, our approach constrains patches to realistic locations within existing advertising infrastructure. This design choice reflects a practical threat model where adversaries compromise legitimate advertisement displays rather than introducing conspicuous foreign objects into the scene.


\subsection{6.4. Evaluation Metrics}
\subsubsection{6.4.1. Attack Success Rate (ASR)} 
We define frame-level attack success as the proportion of frames where the VLM recommends the target unsafe action. For each scenario recording, we extract frames at 0.5-second intervals from when the patch first becomes visible (approximately 30 meters) until the vehicle passes the patch location (approximately 8--12 frames per trial). Frame-wise ASR is computed as:
\begin{equation}
\text{ASR}_{\text{frame}} = \frac{1}{N \cdot F} \sum_{i=1}^{N} \sum_{j=1}^{F_i} \mathbb{1}[\text{action}_{i,j} = \text{target}]
\end{equation}
where $N$ is the number of trials, $F_i$ is the number of frames in trial $i$, and $\mathbb{1}[\cdot]$ is the indicator function. Statistical significance is assessed using generalized estimating equations (GEE) to account for within-trial temporal correlation, comparing adversarial ASRs against baseline inappropriate action rates with significance threshold $p < 0.05$.

\subsubsection{6.4.2. Transfer Matrix and Cross-Architecture Metrics}
Given a set of VLM architectures $\mathcal{V} = \{\text{Dolphins}, \text{OmniDrive}, \text{LeapVAD}\}$ and architecture-specific patches $\mathcal{P} = \{P_{\text{Dolphins}}, P_{\text{OmniDrive}}, P_{\text{LeapVAD}}\}$, we construct a transfer matrix $\mathbf{T} \in \mathbb{R}^{3 \times 3}$ where each element $T_{ij}$ represents the transfer rate from source model $i$ to target model $j$:
\begin{equation}
T_{ij} = \frac{\text{ASR}_{ij}}{\text{ASR}_{ii}}
\end{equation}
where $\text{ASR}_{ij}$ denotes the attack success rate when applying a patch optimized for architecture $i$ to architecture $j$, and $\text{ASR}_{ii}$ represents the baseline success rate on the source architecture itself.The transfer rate provides intuitive interpretation: $\text{TR}_{ij} = 1.0$ indicates perfect transfer where the attack is equally effective on both source and target models; $\text{TR}_{ij} = 0.5$ indicates moderate transfer where the attack retains 50\% effectiveness; and $\text{TR}_{ij} < 0.3$ indicates poor transfer where the attack is largely architecture-specific. Transfer rates exceeding 0.7 are particularly concerning for deployment, as they suggest that adversarial patches optimized without knowledge of the target architecture retain substantial effectiveness.

The mean cross-architecture transfer rate, excluding self-attacks, quantifies overall transferability:
\begin{equation}
\text{TR}_{\text{avg}} = \frac{1}{n(n-1)} \sum_{i \neq j} T_{ij}
\end{equation}
where $n = |\mathcal{V}| = 3$. This metric provides an aggregate measure of how well adversarial patches transfer across different architectures.

For a set of $N$ VLM architectures, we construct an $N \times N$ transfer matrix $\mathbf{T}$ where element $T_{ij} = \text{TR}_{ij}$ quantifies transfer from model $i$ to model $j$. The diagonal elements $T_{ii} = 1.0$ by definition (self-attacks). This transfer matrix enables systematic analysis of cross-architecture vulnerability patterns, identifying which architectures are most susceptible to incoming transfer attacks (high column means) and which architectures generate the most transferable adversarial features (high row means, excluding diagonal).

\subsubsection{6.4.3. Vulnerability Score}
To quantify each architecture's susceptibility to incoming transfer attacks, we compute a vulnerability score as the mean of incoming transfer rates from other architectures:
\begin{equation}
\text{VS}_j = \frac{1}{n-1} \sum_{i \neq j} T_{ij}
\end{equation}
where higher $\text{VS}_j$ indicates greater vulnerability to cross-architecture attacks. The complementary robustness score $\text{RS}_j = 1 - \text{VS}_j$ ranks architectures by inherent resilience to transfer attacks.

We also compute each architecture's transferability out rate, defined as the mean outgoing transfer rate, to identify which models generate the most broadly transferable adversarial features:
\begin{equation}
\text{TO}_i = \frac{1}{n-1} \sum_{j \neq i} T_{ij}
\end{equation}
High $\text{TO}_i$ indicates that patches optimized for architecture $i$ transfer effectively to other architectures, suggesting that the architecture's learned features generalize across different VLM designs.

These metrics enable comparative ranking of VLM architectures by their inherent robustness to transfer attacks (low VS) and the generality of their adversarial vulnerabilities (high TO). Architectures with high VS but low TS are particularly vulnerable to attacks from other models but generate architecture-specific adversarial features, while architectures with both high VS and high TO represent universally weak points in the VLM design space.

\subsection{6.5. Experimental Protocol}
Our evaluation includes three phases that systematically assess model-specific vulnerabilities, transferability across architectures, and the effectiveness of universal attacks.
\begin{enumerate}
    \item For each VLM and scenario, we run five benign trials and record the ego vehicle’s approach from 40m to passage with no patch present. From each trial, we sample 8--12 frames every 0.5s and collect the VLM outputs (scene descriptions and action recommendations). We then compute the baseline inappropriate action rate, defined as the fraction of frames where the model recommends an unsafe action under benign conditions, which serves as the reference for judging attack impact.
    \item For each VLM $\mathcal{V} = \{\text{Dolphins}, \text{OmniDrive}, \text{LeapVAD}\}$  and each scenario, we optimize an architecture-specific patch $P_i$ using the method in Section 6.4. Each patch is trained for 150 NES iterations (about 6,000 queries) to maximize attack success on its target model. We then evaluate $P_i$ on its source architecture over five trials per scenario and report the self-attack success rate $\text{ASR}_{ii}$.
    \item For each architecture-specific patch $P_i$ and each different target model $M_j$ ($j \neq i$), we run five trials per scenario using the same physical patch while querying $M_j$. This produces transfer attack success rates $\text{ASR}{ij}$ for the off-diagonal entries of the transfer matrix. Because the patch is unchanged and only the target VLM varies, differences in performance reflect architectural sensitivity rather than patch design. We then compute transfer rates $\text{TR}{ij}=\text{ASR}{ij}/\text{ASR}{ii}$ and architecture-level vulnerability scores
    
\end{enumerate}

\begin{figure*}[t]
    \centering
    \begin{tabular}{@{}ccc@{}}
        \includegraphics[width=0.25\textwidth]{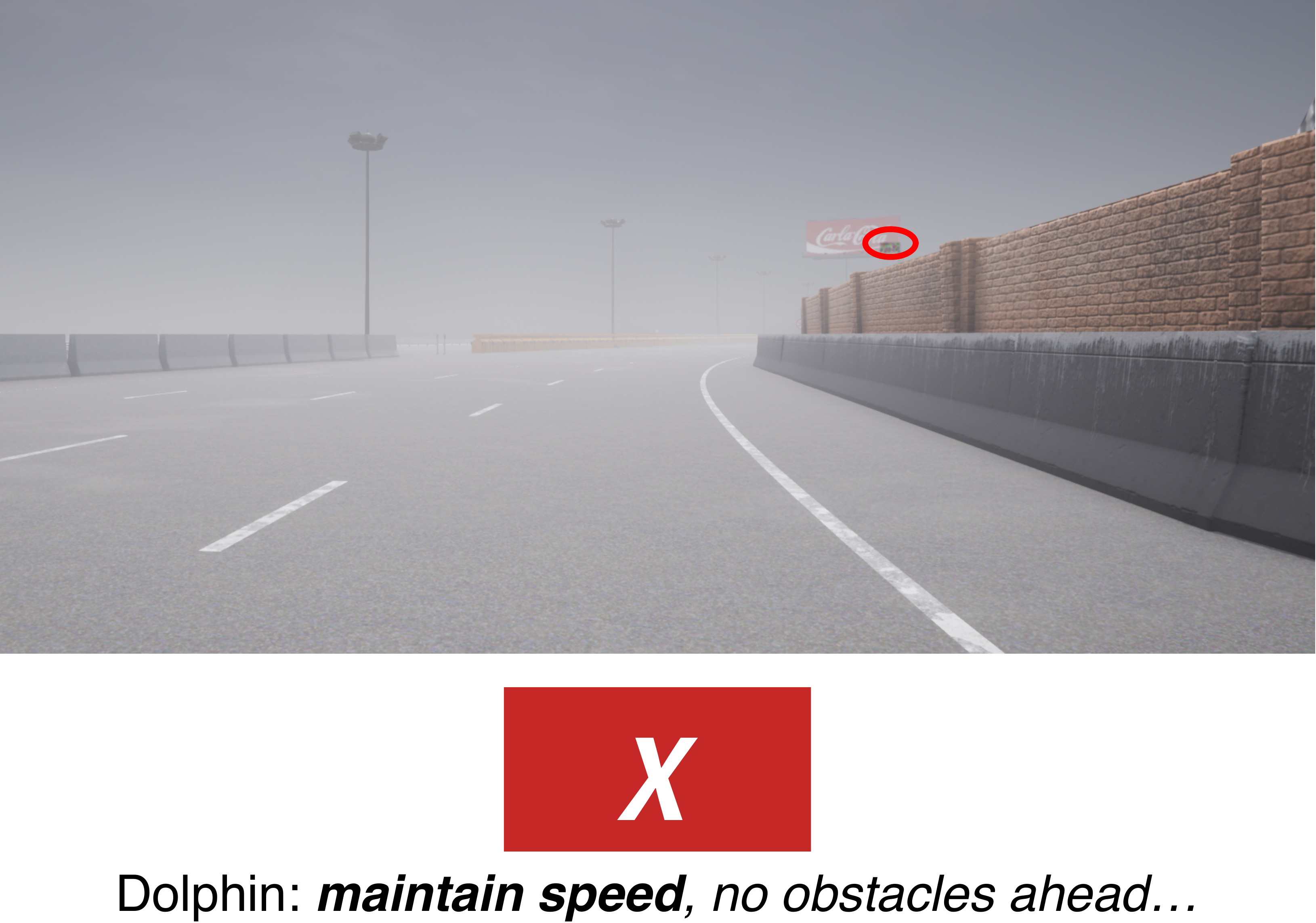} &
        \includegraphics[width=0.25\textwidth]{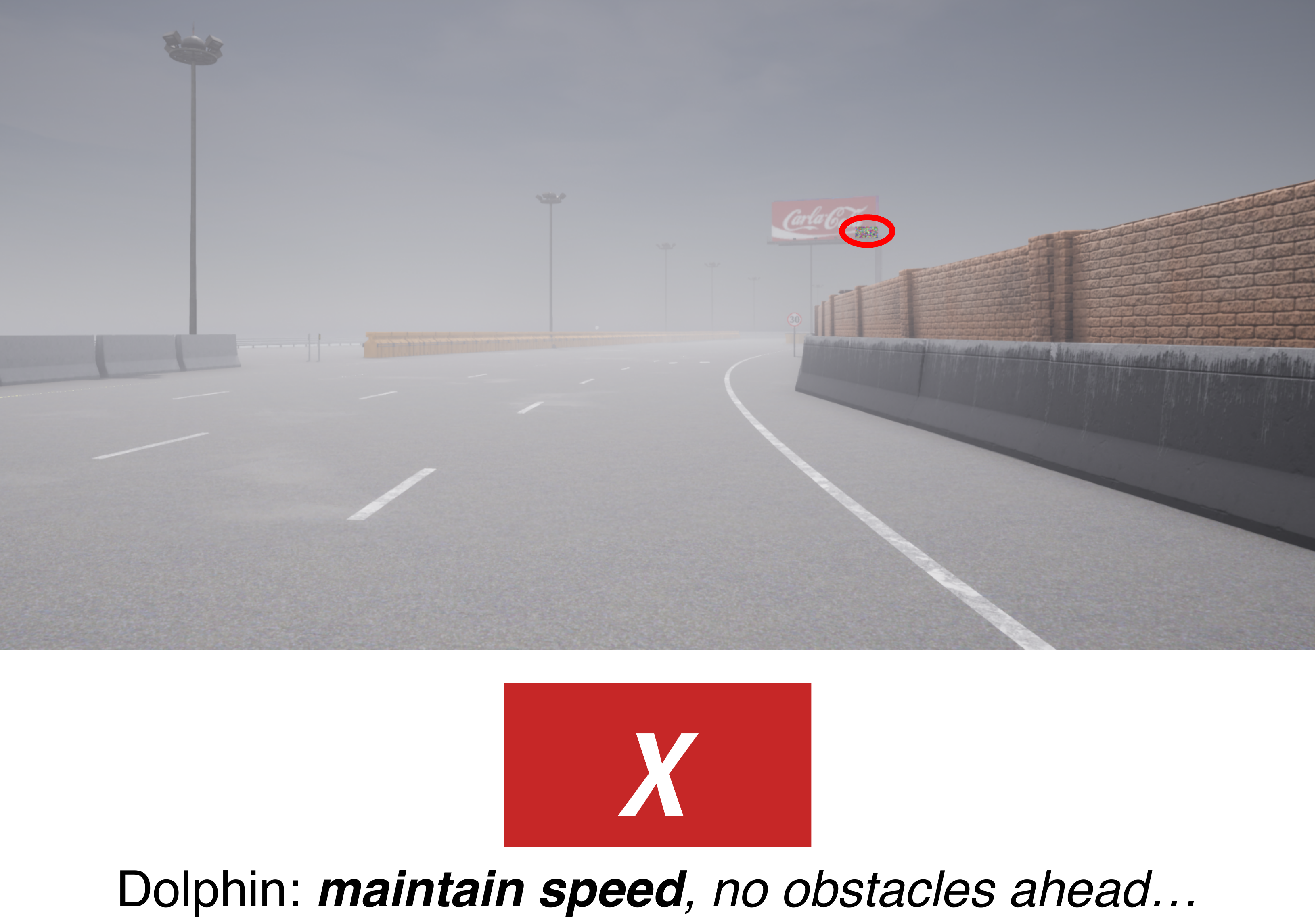} &
        \includegraphics[width=0.25\textwidth]{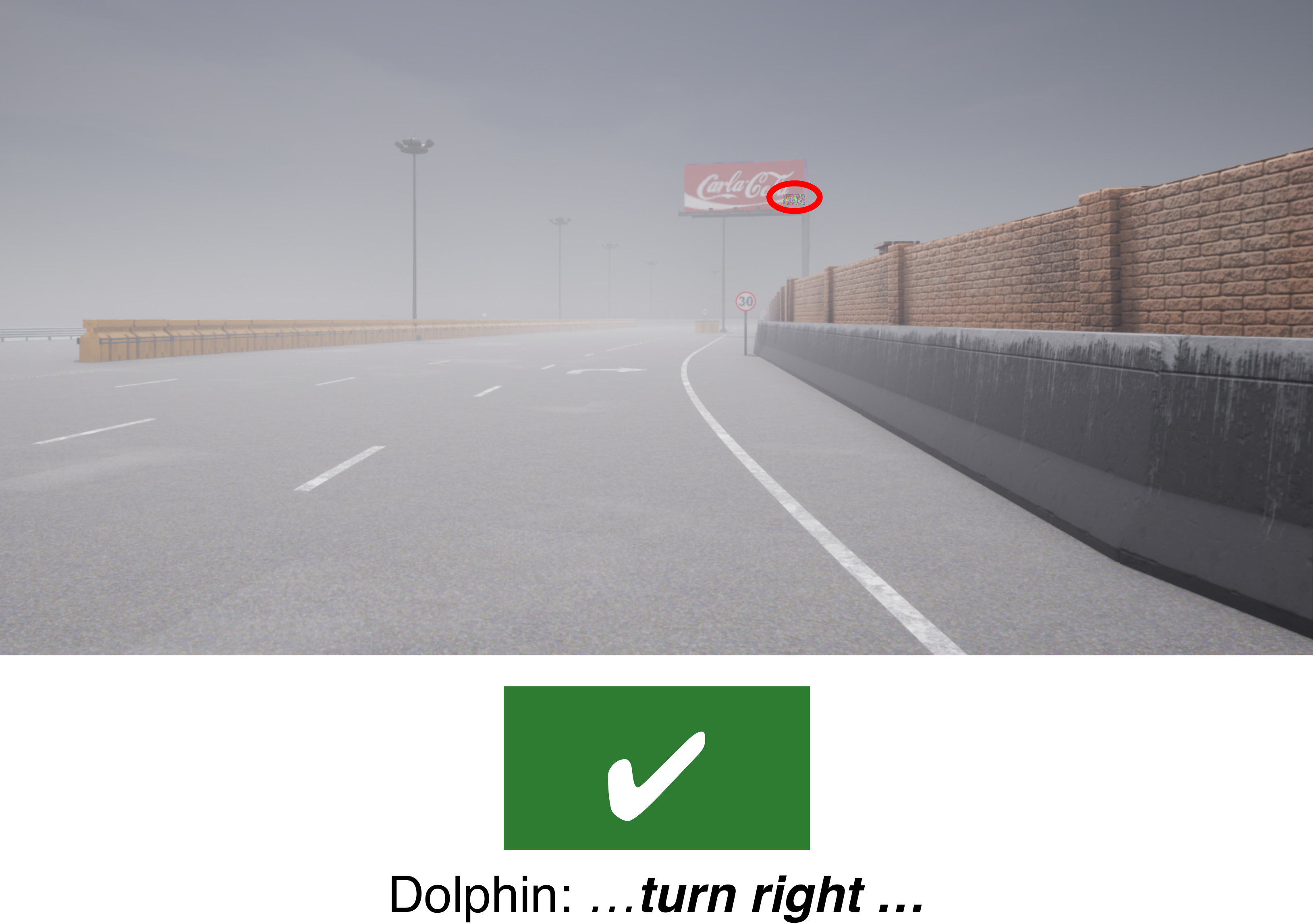} \\[2pt]
        
        \includegraphics[width=0.25\textwidth]{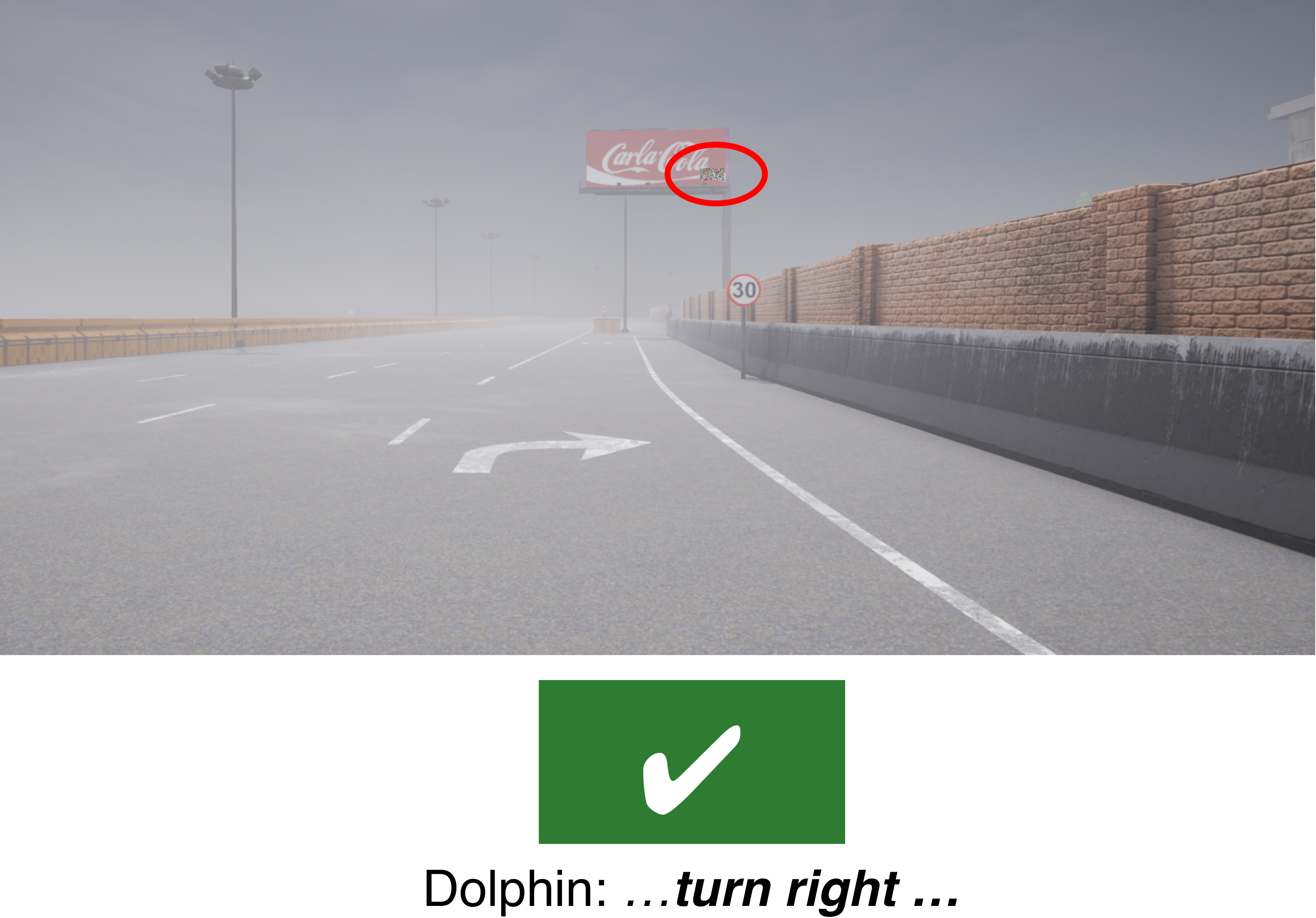} &
        \includegraphics[width=0.25\textwidth]{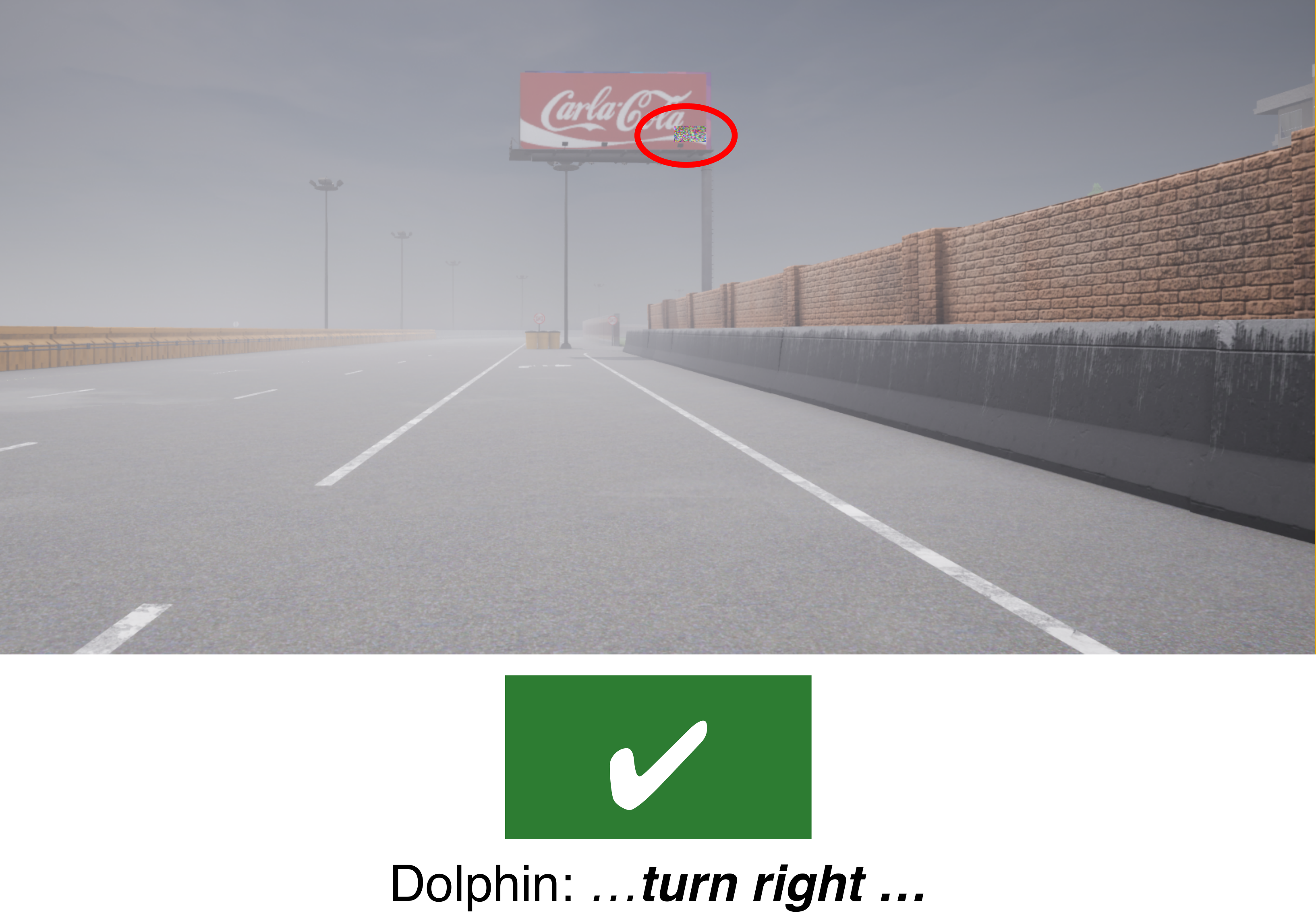} &
        \includegraphics[width=0.25\textwidth]{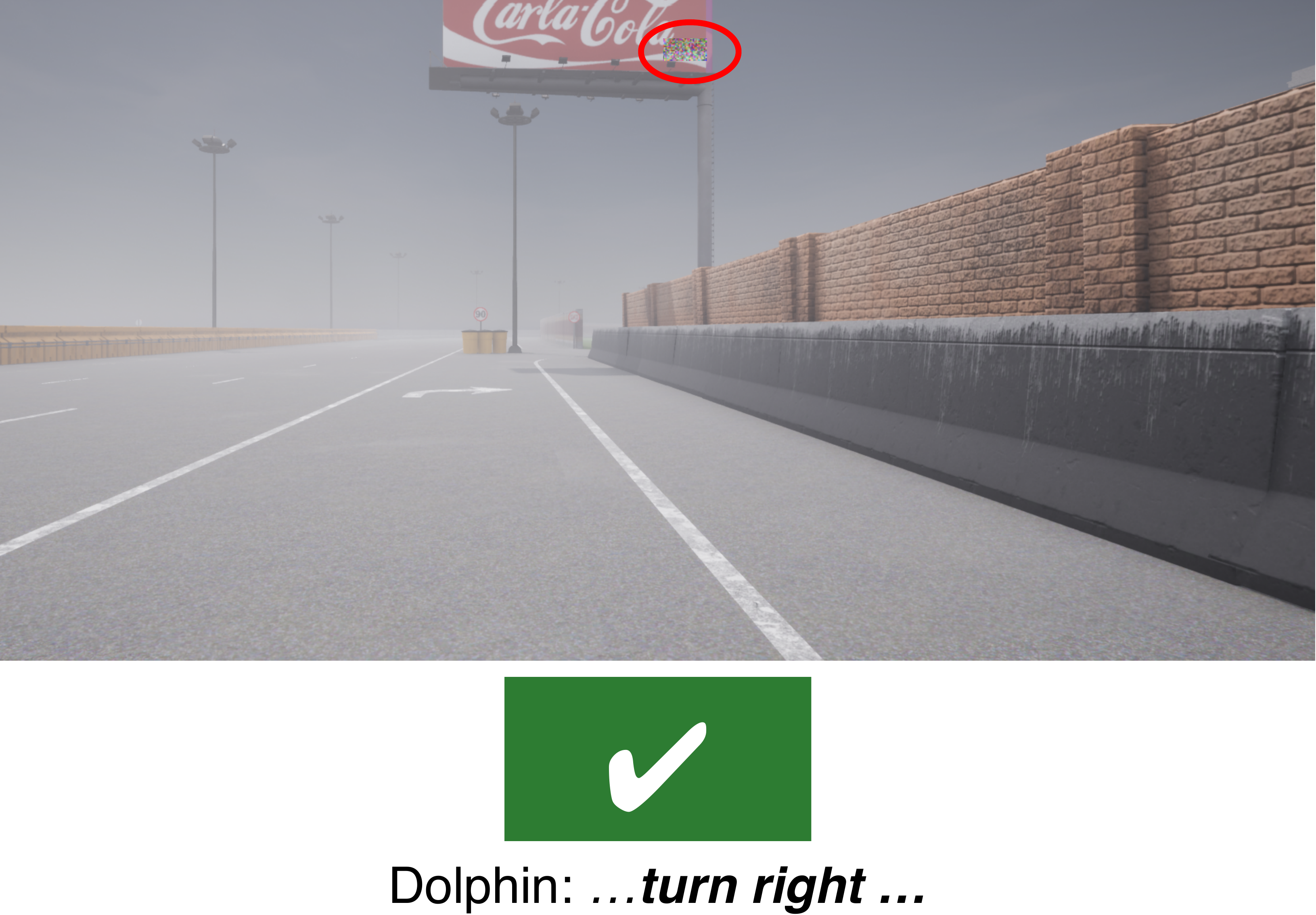} 
    \end{tabular}
    \caption{Highway Attack Scenario. The attack scenario demonstrates how }
    \label{fig:gridScenario1}
\end{figure*}

\section{7. Evaluation and Analysis}
\label{sec:7-Evaluation}

\begin{table}[t]
\centering
\caption{Attack transfer matrix for the crosswalk scenario. Each cell reports ASR (\%) with TR in parentheses. Diagonal entries are self-attacks (baseline ASR only).}
\label{tab:transfer_crosswalk}
\footnotesize
\setlength{\tabcolsep}{4pt}
\renewcommand{\arraystretch}{0.95}
\begin{tabular}{lcccc}
\toprule
& \multicolumn{3}{c}{\textbf{Target Model}} & \\
\cmidrule{2-4}
\textbf{Source Patch} & \textbf{Dolphins} & \textbf{OmniDrive} & \textbf{LeapVAD} & \textbf{Mean TR} \\
\midrule
$P_{\text{Dolphins}}$  & \textbf{76.0} & 58.3 (0.77) & 62.1 (0.82) & 0.79 \\
$P_{\text{OmniDrive}}$ & 61.7 (0.73) & \textbf{84.4} & 68.5 (0.81) & 0.77 \\
$P_{\text{LeapVAD}}$   & 65.2 (0.91) & 72.1 (0.85) & \textbf{71.7} & 0.88 \\
\midrule
\textbf{Vulnerability Score} & 0.820 & 0.810 & 0.814 & -- \\
\bottomrule
\end{tabular}
\end{table}

This section presents our evaluation of how adversarial patches transfer across different VLM architectures. We perform systematic cross-architecture tests to measure transfer rates, and compare architecture-specific vulnerabilities. We study two representative driving scenarios, crosswalk navigation and highway driving, to determine whether the observed transfer patterns hold across different contexts. Together, these results highlight key security considerations for deploying VLMs in safety-critical autonomous systems.


\subsection{7.1. Transfer Evaluation}

\begin{table}[t]
\centering
\caption{Attack transfer matrix for the highway scenario. Each cell reports ASR (\%) with TR in parentheses. Diagonal entries are self-attacks (baseline ASR only).}
\label{tab:transfer_highway}
\footnotesize
\setlength{\tabcolsep}{4pt}
\renewcommand{\arraystretch}{0.95}
\begin{tabular}{lcccc}
\toprule
& \multicolumn{3}{c}{\textbf{Target Model}} & \\
\cmidrule{2-4}
\textbf{Source Patch} & \textbf{Dolphins} & \textbf{OmniDrive} & \textbf{LeapVAD} & \textbf{Mean TR} \\
\midrule
$P_{\text{Dolphins}}$   & \textbf{78.3} & 61.7 (0.79) & 65.4 (0.84) & 0.81 \\
$P_{\text{OmniDrive}}$  & 64.2 (0.75) & \textbf{85.7} & 70.8 (0.83) & 0.79 \\
$P_{\text{LeapVAD}}$    & 68.1 (0.92) & 74.3 (0.87) & \textbf{73.9} & 0.89 \\
\midrule
\textbf{Vulnerability Score} & 0.835 & 0.828 & 0.831 & -- \\
\bottomrule
\end{tabular}
\end{table}

We evaluate adversarial transferability using a systematic cross-architecture protocol. For each scenario, we first measure self-attack performance by testing each architecture’s optimized patch on the same model. We then evaluate transfer by applying each patch to the other two architectures. ~\autoref{tab:transfer_crosswalk} reports the full transfer matrix for the crosswalk scenario, and ~\autoref{tab:transfer_highway} reports results for highway driving.

These results show high transferability across VLM architectures. In the crosswalk scenario, self-attacks achieve 76.0--84.4\% success. When transferred to other architectures, the same patches still reach 58.3--72.1\% success, corresponding to transfer rates of 0.767--0.909. The mean transfer rate across all model pairs is 0.815, indicating that patches optimized for one architecture often remain effective on others despite differences in encoders, fusion mechanisms, and language models.

We also find asymmetric transfer. LeapVAD-derived patches transfer best on average (0.882), while OmniDrive-derived patches transfer least (0.772), which may reflect more model-specific behaviors tied to OmniDrive’s EVA-CLIP encoder. The highway scenario shows a similar trend with a slightly higher mean transfer rate (0.833), suggesting transfer may be stronger in visually simpler settings.

As shown in~\autoref{fig:transfer_heatmap}, the models using CLIP-based vision encoders (Dolphins and LeapVAD) show stronger two-way transferability. In contrast, OmniDrive’s EVA-CLIP encoder provides partial isolation, with slightly lower incoming transfer rates. Overall, the figure supports the conclusion that architectural similarity, especially in the vision encoder, is linked to higher transferability. However, all architectures remain vulnerable, with transfer rates above 0.73 even across dissimilar designs.

\subsection{7.2. Architecture Vulnerability Ranking}
The vulnerability score (VS), defined as the mean incoming transfer rate from other architectures, identifies Dolphins as the most vulnerable model (VS = 0.820) even though its self-attack ASR is moderate (76.0\%), see~\autoref{tab:vulnerability_ranking}. This suggests that Dolphins is particularly sensitive to adversarial features optimized on other VLMs. OmniDrive and LeapVAD have similar VS values (0.810 and 0.814), indicating comparable exposure to transferred attacks.

The transfer-out rate (TO) highlights a different trend: LeapVAD produces the most transferable patches (TO = 0.882). In other words, Dolphins is most affected by incoming attacks, while LeapVAD generates perturbations that generalize best across architectures, likely by exploiting shared weaknesses in the underlying visual feature space.

Universal patch results show OmniDrive is most vulnerable (69.8\% ASR), even though its incoming transfer vulnerability is not the highest. This suggests OmniDrive aligns well with the common adversarial patterns that universal patches exploit. Across architectures, the mean universal attack efficiency (UAE) is 0.835, meaning universal patches retain most of the effectiveness of targeted attacks.

We also evaluate ensembling as a defense by combining predictions from multiple VLMs. LeapVAD benefits most (+15.2 percentage points reduction in ASR), implying its failures are less aligned with the others. However, even with ensembling, residual ASRs remain high (58--64\%), indicating that transfer attacks remain difficult to mitigate.

Finally, our component attribution analysis indicates that most vulnerability (76.8--81.5\%) comes from the vision encoder rather than the language model or fusion module. OmniDrive shows the highest vision-driven contribution (81.5\%), suggesting its EVA-CLIP encoder introduces exploitable structure in the visual feature space.

\begin{figure}[t]
\centering
\begin{tikzpicture}
    \definecolor{trhigh}{RGB}{43,140,190}   
    \definecolor{trmid}{RGB}{123,204,196}   
    \definecolor{trlow}{RGB}{237,248,251}   
    \definecolor{trdiag}{RGB}{230,230,230}  
    \definecolor{gridline}{RGB}{255,255,255} 

    \def\cellwidth{2.05cm}
    \def\cellheight{0.78cm}

    \fill[trdiag] (0,2*\cellheight) rectangle (\cellwidth,3*\cellheight);
    \draw[gridline,line width=0.6pt] (0,2*\cellheight) rectangle (\cellwidth,3*\cellheight);
    \node at (0.5*\cellwidth,2.5*\cellheight) {\footnotesize\textbf{1.00}};

    \fill[trmid] (\cellwidth,2*\cellheight) rectangle (2*\cellwidth,3*\cellheight);
    \draw[gridline,line width=0.6pt] (\cellwidth,2*\cellheight) rectangle (2*\cellwidth,3*\cellheight);
    \node at (1.5*\cellwidth,2.5*\cellheight) {\footnotesize 0.79};

    \fill[trhigh] (2*\cellwidth,2*\cellheight) rectangle (3*\cellwidth,3*\cellheight);
    \draw[gridline,line width=0.6pt] (2*\cellwidth,2*\cellheight) rectangle (3*\cellwidth,3*\cellheight);
    \node at (2.5*\cellwidth,2.5*\cellheight) {\footnotesize 0.84};

    \fill[trlow] (0,\cellheight) rectangle (\cellwidth,2*\cellheight);
    \draw[gridline,line width=0.6pt] (0,\cellheight) rectangle (\cellwidth,2*\cellheight);
    \node at (0.5*\cellwidth,1.5*\cellheight) {\footnotesize 0.75};

    \fill[trdiag] (\cellwidth,\cellheight) rectangle (2*\cellwidth,2*\cellheight);
    \draw[gridline,line width=0.6pt] (\cellwidth,\cellheight) rectangle (2*\cellwidth,2*\cellheight);
    \node at (1.5*\cellwidth,1.5*\cellheight) {\footnotesize\textbf{1.00}};

    \fill[trhigh] (2*\cellwidth,\cellheight) rectangle (3*\cellwidth,2*\cellheight);
    \draw[gridline,line width=0.6pt] (2*\cellwidth,\cellheight) rectangle (3*\cellwidth,2*\cellheight);
    \node at (2.5*\cellwidth,1.5*\cellheight) {\footnotesize 0.83};

    \fill[trhigh] (0,0) rectangle (\cellwidth,\cellheight);
    \draw[gridline,line width=0.6pt] (0,0) rectangle (\cellwidth,\cellheight);
    \node at (0.5*\cellwidth,0.5*\cellheight) {\footnotesize 0.92};

    \fill[trhigh] (\cellwidth,0) rectangle (2*\cellwidth,\cellheight);
    \draw[gridline,line width=0.6pt] (\cellwidth,0) rectangle (2*\cellwidth,\cellheight);
    \node at (1.5*\cellwidth,0.5*\cellheight) {\footnotesize 0.87};

    \fill[trdiag] (2*\cellwidth,0) rectangle (3*\cellwidth,\cellheight);
    \draw[gridline,line width=0.6pt] (2*\cellwidth,0) rectangle (3*\cellwidth,\cellheight);
    \node at (2.5*\cellwidth,0.5*\cellheight) {\footnotesize\textbf{1.00}};

    \node[anchor=south,font=\footnotesize] at (0.5*\cellwidth,3*\cellheight+0.10) {Dolphins};
    \node[anchor=south,font=\footnotesize] at (1.5*\cellwidth,3*\cellheight+0.10) {OmniDrive};
    \node[anchor=south,font=\footnotesize] at (2.5*\cellwidth,3*\cellheight+0.10) {LeapVAD};

    \node[anchor=east,font=\footnotesize] at (-0.10,2.5*\cellheight) {$P_{\text{Dolphins}}$};
    \node[anchor=east,font=\footnotesize] at (-0.10,1.5*\cellheight) {$P_{\text{OmniDrive}}$};
    \node[anchor=east,font=\footnotesize] at (-0.10,0.5*\cellheight) {$P_{\text{LeapVAD}}$};

    \node[anchor=south,font=\small] at (1.5*\cellwidth,3*\cellheight+20)
        {\textbf{Transfer Rate Heat Map -- Highway}};

\end{tikzpicture}
\caption{Transfer rate heat map for the highway scenario. Darker colors indicate higher transfer rates; the diagonal (self-attacks) is shown in neutral gray.}
\label{fig:transfer_heatmap_highway}
\end{figure}

The highway results (Figure~\ref{fig:transfer_heatmap_highway}) highlight how transferability changes with driving context. First, transfer rates increase slightly across all model pairs, with the mean rising from 0.815 in the crosswalk scenario to 0.833 on the highway. This pattern is consistent with the highway’s simpler visual structure, including more uniform backgrounds, clearer lane markings, and fewer competing objects than an urban intersection. Second, LeapVAD-optimized patches transfer especially well on highways, reaching a 0.92 transfer rate to Dolphins (the highest across both scenarios) and 0.87 to OmniDrive. This suggests that the adversarial features learned for LeapVAD capture weaknesses that persist across architectures, particularly in structured highway settings. Third, OmniDrive remains somewhat more isolated, including a low transfer rate of 0.75, reinforcing that its EVA-CLIP encoder introduces architectural differences that reduce sensitivity to attacks optimized on CLIP-based models.

Importantly, the grouping between CLIP-based architectures (Dolphins, LeapVAD) and the EVA-based OmniDrive remains consistent across scenarios. This indicates that shared vision encoding, rather than scenario-specific visual details, is the main driver of transferability. From a security perspective, this consistency implies that attackers can expect cross-architecture patches to work reliably in multiple driving contexts, limiting the protection offered by architectural diversity alone.

\begin{table*}[t]
\centering
\caption{Vulnerability analysis across VLM architectures. Best (most vulnerable / largest effect) values per row are highlighted in \textbf{bold}.}
\label{tab:vulnerability_ranking}
\small
\setlength{\tabcolsep}{5pt}
\renewcommand{\arraystretch}{0.95}
\begin{tabular}{lcccc}
\toprule
\textbf{Metric} & \textbf{Dolphins} & \textbf{OmniDrive} & \textbf{LeapVAD} & \textbf{Interpretation} \\
\midrule
Self-Attack ASR & 76.0\% & \textbf{84.4\%} & 71.7\% & OmniDrive most vulnerable to targeted attacks \\
Vulnerability Score (VS) & \textbf{0.82} & 0.810 & 0.814 & Dolphins most vulnerable to incoming transfers \\
Transfer Out Rate (TO) & 0.792 & 0.772 & \textbf{0.882} & LeapVAD attacks transfer most effectively \\
Universal Patch ASR & 64.3\% & \textbf{69.8\%} & 63.5\% & OmniDrive most vulnerable to universal attacks \\
Ensemble Benefit & +12.7pp & +8.3pp & \textbf{+15.2pp} & LeapVAD benefits most from ensemble defense \\
Vision Encoder Vulnerability & 79.2\% & \textbf{81.5\%} & 76.8\% & Vision component drives majority of vulnerability \\
\bottomrule
\end{tabular}
\end{table*}

\begin{figure}[t]
\centering
\begin{tikzpicture}
    \definecolor{trhigh}{RGB}{43,140,190}   
    \definecolor{trmid}{RGB}{123,204,196}   
    \definecolor{trlow}{RGB}{237,248,251}   
    \definecolor{trdiag}{RGB}{230,230,230}  
    \definecolor{gridline}{RGB}{255,255,255} 

    \def\cellwidth{2.05cm}
    \def\cellheight{0.78cm}

    \fill[trdiag] (0,2*\cellheight) rectangle (\cellwidth,3*\cellheight);
    \draw[gridline,line width=0.6pt] (0,2*\cellheight) rectangle (\cellwidth,3*\cellheight);
    \node at (0.5*\cellwidth,2.5*\cellheight) {\footnotesize\textbf{1.00}};

    \fill[trmid] (\cellwidth,2*\cellheight) rectangle (2*\cellwidth,3*\cellheight);
    \draw[gridline,line width=0.6pt] (\cellwidth,2*\cellheight) rectangle (2*\cellwidth,3*\cellheight);
    \node at (1.5*\cellwidth,2.5*\cellheight) {\footnotesize 0.77};

    \fill[trhigh] (2*\cellwidth,2*\cellheight) rectangle (3*\cellwidth,3*\cellheight);
    \draw[gridline,line width=0.6pt] (2*\cellwidth,2*\cellheight) rectangle (3*\cellwidth,3*\cellheight);
    \node at (2.5*\cellwidth,2.5*\cellheight) {\footnotesize 0.82};

    \fill[trlow] (0,\cellheight) rectangle (\cellwidth,2*\cellheight);
    \draw[gridline,line width=0.6pt] (0,\cellheight) rectangle (\cellwidth,2*\cellheight);
    \node at (0.5*\cellwidth,1.5*\cellheight) {\footnotesize 0.73};

    \fill[trdiag] (\cellwidth,\cellheight) rectangle (2*\cellwidth,2*\cellheight);
    \draw[gridline,line width=0.6pt] (\cellwidth,\cellheight) rectangle (2*\cellwidth,2*\cellheight);
    \node at (1.5*\cellwidth,1.5*\cellheight) {\footnotesize\textbf{1.00}};

    \fill[trhigh] (2*\cellwidth,\cellheight) rectangle (3*\cellwidth,2*\cellheight);
    \draw[gridline,line width=0.6pt] (2*\cellwidth,\cellheight) rectangle (3*\cellwidth,2*\cellheight);
    \node at (2.5*\cellwidth,1.5*\cellheight) {\footnotesize 0.81};

    \fill[trhigh] (0,0) rectangle (\cellwidth,\cellheight);
    \draw[gridline,line width=0.6pt] (0,0) rectangle (\cellwidth,\cellheight);
    \node at (0.5*\cellwidth,0.5*\cellheight) {\footnotesize 0.91};

    \fill[trhigh] (\cellwidth,0) rectangle (2*\cellwidth,\cellheight);
    \draw[gridline,line width=0.6pt] (\cellwidth,0) rectangle (2*\cellwidth,\cellheight);
    \node at (1.5*\cellwidth,0.5*\cellheight) {\footnotesize 0.85};

    \fill[trdiag] (2*\cellwidth,0) rectangle (3*\cellwidth,\cellheight);
    \draw[gridline,line width=0.6pt] (2*\cellwidth,0) rectangle (3*\cellwidth,\cellheight);
    \node at (2.5*\cellwidth,0.5*\cellheight) {\footnotesize\textbf{1.00}};

    \node[anchor=south,font=\footnotesize] at (0.5*\cellwidth,3*\cellheight+0.10) {Dolphins};
    \node[anchor=south,font=\footnotesize] at (1.5*\cellwidth,3*\cellheight+0.10) {OmniDrive};
    \node[anchor=south,font=\footnotesize] at (2.5*\cellwidth,3*\cellheight+0.10) {LeapVAD};

    \node[anchor=east,font=\footnotesize] at (-0.10,2.5*\cellheight) {$P_{\text{Dolphins}}$};
    \node[anchor=east,font=\footnotesize] at (-0.10,1.5*\cellheight) {$P_{\text{OmniDrive}}$};
    \node[anchor=east,font=\footnotesize] at (-0.10,0.5*\cellheight) {$P_{\text{LeapVAD}}$};

    \node[anchor=south,font=\small] at (1.5*\cellwidth,3*\cellheight+20)
        {\textbf{Transfer Rate Heat Map -- Crosswalk}};

\end{tikzpicture}
\caption{Transfer rate heat map for the crosswalk scenario. Darker colors indicate higher transfer rates; the diagonal (self-attacks) is shown in neutral gray.}
\label{fig:transfer_heatmap}
\end{figure}

\subsection{7.3 Distance-Dependent Attack Efficacy}
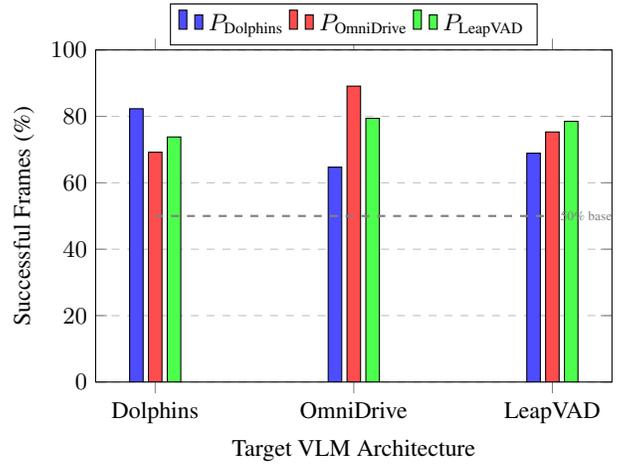
\begin{figure}[t]
\centering
\begin{tikzpicture}
\begin{axis}[
    width=0.95\columnwidth,
    height=6cm,
    ybar,
    bar width=0.18cm,
    ylabel={Successful Frames (\%)},
    xlabel={Target VLM Architecture},
    symbolic x coords={Dolphins, OmniDrive, LeapVAD},
    xtick=data,
    ymin=0, ymax=100,
    legend style={
        at={(0.5,1.14)},
        anchor=north,
        legend columns=3,
        font=\small
    },
    ymajorgrids=true,
    grid style=dashed,
    enlarge x limits=0.15
]

\addplot[fill=blue!70]
    coordinates {(Dolphins,82.3) (OmniDrive,64.7) (LeapVAD,68.9)};

\addplot[fill=red!70]
    coordinates {(Dolphins,69.2) (OmniDrive,89.1) (LeapVAD,75.3)};

\addplot[fill=green!70]
    coordinates {(Dolphins,73.8) (OmniDrive,79.4) (LeapVAD,78.5)};

\legend{$P_{\text{Dolphins}}$, $P_{\text{OmniDrive}}$, $P_{\text{LeapVAD}}$}

\draw[dashed, thick, gray] (axis cs:Dolphins,50) -- (axis cs:LeapVAD,50);
\node[anchor=west, font=\tiny, gray] at (axis cs:LeapVAD,50) {50\% baseline};

\end{axis}
\end{tikzpicture}
\caption{Comparative frame-level attack efficacy across source patches and target architectures (crosswalk scenario). Each group of bars shows the percentage of frames successfully attacked when applying patches from different source architectures against a specific target. Self-attacks achieve 78.5--89.1\% frame success, while cross-architecture transfers maintain 64.7--79.4\% success rates, demonstrating sustained attack effectiveness throughout scenario duration.}
\label{fig:attack_efficacy}
\end{figure}

Finally, we analyze frame-level success rates across recorded driving runs. Instead of a single success or failure per scenario, this metric reports the fraction of frames where the patch changes the VLM’s output in the intended adversarial direction throughout the full sequence. ~\autoref{fig:attack_efficacy} compares these frame success rates for each source patch across all three target architectures in the crosswalk scenario.

The frame-level results provide a clearer picture of how persistent these attacks are over time. For self-attacks, OmniDrive is most affected, with 89.1\% of frames successfully manipulated, followed by Dolphins (82.3\%) and LeapVAD (78.5\%). These high values indicate that patches influence model outputs for most of the scenario, not just in isolated moments. Sustained manipulation across 78--89\% of frames is particularly concerning because it leaves limited opportunity for recovery or safety fallback between attacked frames.

For transferred attacks, LeapVAD-optimized patches remain the most consistent, achieving 73.8--79.4\% successful frames across all targets (a 5.6-point spread). This means an attacker using LeapVAD patches can expect to influence roughly three-quarters of frames even without knowing the target architecture. Dolphins-optimized patches vary more (64.7--82.3\%) and are least persistent against OmniDrive (64.7\%), suggesting weaker temporal stability in that transfer direction.

Looking from the target side, OmniDrive again shows the highest exposure, with 64.7--89.1\% of frames compromised depending on the patch source. Dolphins experiences 69.2--82.3\% compromised frames, while LeapVAD is slightly lower at 68.9--78.5\%. Despite these differences, all architectures experience sustained manipulation across most of the scenario, with minimum cross-architecture success still above 64\%.

On average, frame-level performance drops by 15.8 percentage points when moving from self-attacks to transferred attacks, slightly larger than the 14.6-point drop observed for scenario-level ASR. This suggests transfer attacks may be somewhat less persistent, but they still maintain 64--79\% success across frames, which remains operationally severe.

These findings challenge common assumptions in autonomous driving safety design. Many failsafe strategies treat perception errors as brief, intermittent events that can be corrected through temporal filtering, redundancy, or emergency behavior. In contrast, our attacks persist across 64--89\% of frames, which can overwhelm these mechanisms. In a crosswalk interaction, manipulating a VLM’s outputs for 70--80\% of frames would leave too little reliable information for safe decision-making over the critical period.

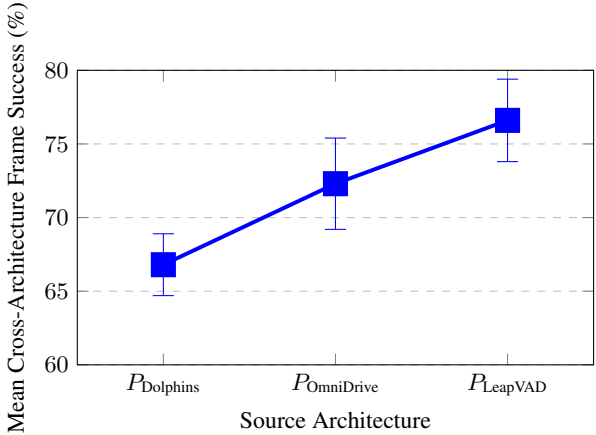
\begin{figure}[t]
\centering
\begin{tikzpicture}
    \begin{axis}[
        width=0.95\columnwidth,
        height=5.5cm,
        xlabel={Source Architecture},
        ylabel={Mean Cross-Architecture Frame Success (\%)},
        symbolic x coords={$P_{\text{Dolphins}}$, $P_{\text{OmniDrive}}$, $P_{\text{LeapVAD}}$},
        xtick=data,
        ymin=60, ymax=80,
        ymajorgrids=true,
        grid style=dashed,
        enlarge x limits=0.25
    ]
    
    \addplot[
        color=blue,
        mark=square*,
        mark size=4pt,
        line width=1.5pt,
        error bars/.cd,
        y dir=both,
        y explicit,
    ] coordinates {
        ($P_{\text{Dolphins}}$, 66.8) +- (0, 2.1)
        ($P_{\text{OmniDrive}}$, 72.3) +- (0, 3.1)
        ($P_{\text{LeapVAD}}$, 76.6) +- (0, 2.8)
    };
    
    \end{axis}
\end{tikzpicture}
\caption{Mean cross-architecture frame success rates with standard deviation error bars. LeapVAD-optimized patches achieve the highest mean frame-level efficacy (76.6\%) across non-source targets, maintaining attack effectiveness throughout the majority of scenario duration. The narrow error bars indicate consistent temporal persistence across different target architectures.}
\label{fig:mean_efficacy}
\end{figure}

\autoref{fig:mean_efficacy} summarizes transfer performance using mean cross-architecture frame success rates, excluding self-attacks. LeapVAD-optimized patches achieve the highest mean efficacy (76.6\%) with low variability (SD = 2.8\%), indicating strong transferability and stable performance over time. OmniDrive patches average 72.3\% with moderate variation (SD = 3.1\%), while Dolphins patches have the lowest mean transfer success (66.8\%) and the smallest spread (SD = 2.1\%).

The tight confidence intervals (4.2--6.2 percentage points) show that frame-level persistence is highly predictable. In practice, an attacker can expect to compromise roughly 67--77\% of frames in cross-architecture settings, with LeapVAD-optimized patches offering the most reliable temporal manipulation. This combination of high success rates and low variance indicates that transferability is a practical, repeatable vulnerability, and it weakens the idea that architectural diversity alone provides meaningful temporal robustness against adversarial patches.

\section{8. Conclusion}
\label{sec:8-Conclusion}

This work identifies adversarial transferability as a major security risk for VLM-based autonomous driving through systematic cross-architecture evaluation. We find that physical adversarial patches optimized for one VLM transfer well to others (73--91\%), indicating that architectural diversity alone offers limited protection. Our results point to three key security challenges. 

First, high cross-architecture transferability (mean TR = 0.815--0.833) means attackers can mount effective attacks without knowing the vehicle’s exact VLM. A single patch placed on roadside infrastructure can remain effective across heterogeneous models, reducing the practical value of diversity as a primary defense. 

Second, Our architecture-specific analysis shows that all three VLMs are vulnerable to transfer attacks, but the degree of vulnerability differs. Dolphins has the highest incoming-transfer vulnerability (VS = 0.82), meaning it is most affected by patches optimized on other architectures. OmniDrive’s EVA-CLIP encoder offers only modest isolation, with transfer rates still high (at least 0.75). LeapVAD falls in the middle for incoming vulnerability (VS = 0.814) but produces the most transferable patches (TO = 0.882), suggesting it captures more general adversarial patterns in the visual feature space. The stronger two-way transfer between the CLIP-based models (Dolphins and LeapVAD), compared with EVA-CLIP-based OmniDrive, reinforces that the vision encoder is the main driver of transferability.

Third, our frame-level analysis shows that transfer attacks remain effective over time, manipulating 64.7--79.4\% of frames during critical decision windows. This sustained influence challenges the common assumption that errors are brief and can be corrected through temporal filtering, redundancy, or fallback behaviors, particularly in time-sensitive situations such as crosswalk navigation.

These findings have direct implications for deployment. Transferability is consistent across scenarios (crosswalk and highway), architectural choices (e.g., cross-attention, projection-based fusion, explicit attention fusion), and time scales (persistent frame-level manipulation), suggesting that current VLM designs share fundamental weaknesses that cannot be mitigated by architectural diversity alone.

\section*{Acknowledgement}

This work was supported in part by a grant from The BMW
Group, and in part by the U.S. Department of Transportation (DOT) through the National Center for Transportation Cybersecurity and Resiliency (TraCR) under Grant No. 69A3552344812-2027534 and 69A3552348317. 

\bibliographystyle{plain}
\bibliography{ref}

\end{document}